\documentclass[letterpaper, 10 pt, conference]{ieeeconf}  

\IEEEoverridecommandlockouts                              

\usepackage{graphics} 
\usepackage{times} 
\usepackage{amsmath} 
\usepackage{amssymb}  
\usepackage{mathtools,amsfonts,amssymb,bm}
\usepackage{hyperref}
\usepackage[demo]{graphicx}
\usepackage{subcaption}
\usepackage{multirow}
\usepackage{xcolor}
\usepackage{soul}
\usepackage{cite}
\usepackage{float}
\usepackage{algorithm}
\usepackage{algpseudocode}
\usepackage{optidef}
\usepackage{anyfontsize}
\usepackage{booktabs}
 
\usepackage{comment}
\usepackage{microtype}
\usepackage{todonotes}

\title{\LARGE \bf
Physics-Informed Learning for Human Whole-Body Kinematics Prediction via Sparse IMUs
}

\author{Cheng Guo$^{1,2}$, Giuseppe L'Erario$^{1,2}$,  Giulio Romualdi$^{1}$,  Mattia Leonori$^{3}$, \\ 
Marta Lorenzini$^{3}$, Arash Ajoudani$^{3}$, Daniele Pucci$^{1,2}$
\thanks{*This work was supported by the Italian National Institute for Insurance against Accidents at Work (INAIL) ergoCub Project.}
\thanks{$^{1}$Artificial and Mechanical Intelligence, Italian Institute of Technology, Genoa, Italy. 
        email: {\tt\small { firstname.lastname@iit.it}}}%
\thanks{$^{2}$Department of Computer Science, The University of Manchester, Manchester, United Kingdom.}%
\thanks{$^{3}$Human-Robot Interfaces and Interaction, Italian Institute of Technology, Genoa, Italy. 
        email: {\tt\small { firstname.lastname@iit.it}}}%
}

\begin{document}

\maketitle
\thispagestyle{empty}
\pagestyle{empty}

\begin{abstract}

Accurate and physically feasible human motion prediction is crucial for safe and seamless human-robot collaboration. 
While recent advancements in human motion capture enable real-time pose estimation, 
the practical value of many existing approaches is limited by the lack of future predictions and consideration of physical constraints.
Conventional motion prediction schemes rely heavily on past poses, which are not always available in real-world scenarios. 
To address these limitations, we present a physics-informed learning framework that integrates domain knowledge into both training and inference to predict human motion using inertial measurements from only 5 IMUs.
We propose a network that accounts for the spatial characteristics of human movements. 
During training, we incorporate forward and differential kinematics functions as additional loss components to regularize the learned joint predictions. 
At the inference stage, we refine the prediction from the previous iteration to update a joint state buffer, which is used as extra inputs to the network.
Experimental results demonstrate that our approach achieves high accuracy, smooth transitions between motions, and generalizes well to unseen subjects. 
The source code and data are available at \href{https://github.com/ami-iit/paper_guo_2025_iros_neural_kinematics_prediction}{$https://github.com/ami-iit/paper \textunderscore guo \textunderscore 2025 \textunderscore iros \textunderscore human \textunderscore kinematics \textunderscore prediction$}.


\end{abstract}

\section{INTRODUCTION}
\label{sec:introduction}
The popularity of robots has surged in recent years due to their environmental interaction capabilities, particularly through the growing focus on \emph{Human-Robot Collaboration} (HRC), which envisions robots working alongside humans to enhance productivity and efficiency \cite{Ajoudani2018}. 
A vital trait for robots to ensure safe and seamless collaboration is the ability to predict human intentions and future movements. 
With progress in sensing technologies and deep learning algorithms, various methods have been developed to reconstruct human motions from sensor measurements \cite{niu2024method}. 
However, inferring human intentions and predicting future movements from partial observations remains a significant challenge due to the complexity and adaptability of human dynamics. 
This paper proposes a framework that adopts the idea of physics-informed (PI) learning to enable the prediction of physically plausible human whole-body joint kinematics from sparse inertial measurements. 

\begin{figure}[h]
    \centering
    \includegraphics[width=\linewidth]{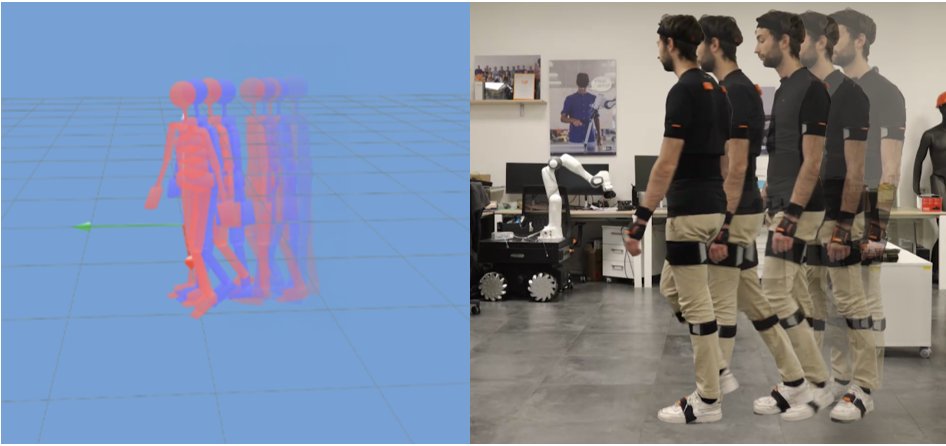}
    \caption{The proposed approach predicts human joint kinematics via 5 IMUs attached to the pelvis, left/right forearms, and left/right lower legs. The blue model is the reconstructed human motion at the current timestamp (corresponding to the human subject on the right), and the red model is the prediction 10 steps ahead (about 167 ms).}
    \label{fig:first_page_fig}
\end{figure}

Traditional \emph{human motion capture} techniques primarily rely on \emph{optical-based} methods, using monocular RGB cameras \cite{habermann2020deepcap} or depth cameras \cite{zheng2021deepmulticap} to reconstruct human poses. While these methods achieve high accuracy, they are constrained by narrow activity spaces and are highly sensitive to environmental factors such as lighting and occlusion.
Consequently, \emph{inertial-based} approaches have gained increasing attention. Commercial motion capture solutions \cite{xsens, noitom} provide precise pose estimation but require a dense placement of numerous inertial measurement units (IMUs), which could be inconvenient for industrial scenarios. Recent efforts have explored motion reconstruction with fewer IMUs \cite{Huang2018, Yi2021, Yi2022, Jiang2022}, but challenges such as long-term drift, resolving ambiguous poses and predicting future motions remain, which are critical for improving efficiency and robustness in HRC settings.

To enable robots to predict human motions, many studies leverage deep learning techniques and frame the task as a \emph{sequence-to-sequence} problem based on past poses \cite{koppula2015anticipating, gui2018teaching, zhang2022pimnet}.
However, 
when modeling humans as highly articulated kinematics chains, motion is typically represented as the evolution of whole-body joint positions. In practice, motion capture systems record limb poses, 
thus requiring an inverse kinematics solution to map these limb configurations to joint positions.
Few works have pursued \emph{End-to-End} approaches using wearable sensors \cite{Katsuhara2019, Yang2021, Zhang2023}, these methods often face restrictions such as short prediction horizons
and complex sensor integration, limiting their real-world applicability. 

More recently, the approach of using \textit{Physics-Informed Neural Networks} (PINNs) has shown considerable potential in solving complex problems in various scientific and engineering fields where data is limited or noisy \cite{Raissi2019}. By embedding physical laws, typically expressed as partial differential equations (PDEs), directly into the neural network architecture, PINNs enable  
more accurate and interpretable modeling of systems with inherent physical behaviors.
Inspired by this, researchers have attempted to integrate Newtonian Dynamics Equations with different deep architectures to predict muscle forces and joint angles from surface electromyogram (sEMG) signals for biomechanical analysis \cite{Zhang2022, Ma2024, Shi2023}. However, these approaches are not directly applicable for robot-related scenarios due to limitations such as signal variability and constrained movement ranges.

To this end, we present a PINN-based model that leverages neural networks alongside physics equations to predict human whole-body kinematics using the inertial data from only 5 IMUs (see Figure \ref{fig:first_page_fig}). 
We represent the human model using the Unified Robot Description Format (URDF), a standard framework for defining rigid-body structures in robotics, 
which allows a structured definition of human motion in the configuration space.
The forward and differential kinematics functions are integrated into the training to guide the network in learning the intrinsic relationships between joint and limb kinematics, reducing reliance on supervision labels and improving generalization.
To enable smoother transitions between motions, we introduce a \textit{joint state buffer} that 
incorporates predictions from the previous iteration as extra inputs for subsequent inference, allowing for recursive updates.
In conclusion, the main contributions are:
\begin{itemize}
  \item 
  We propose a PI learning framework that incorporates a network architecture designed to account for the temporal and spatial characteristics of human movements. 
  \item During training, we formulate forward and differential kinematics functions as additional losses, guiding the network to learn the underlying physical relationships that are not explicitly present in the training data. 
  \item We introduce a joint state buffer to achieve smoother transitions between motions. At inference, the previous prediction is refined to update the buffer.
  \item We demonstrate the performance of the proposed method in both simulation and real-world experiments.
\end{itemize}

The remain of this paper is organized as follows. Section \ref{sec:background} introduces the most related work. Section \ref{sec:methods} provides the details of our approach.
Section \ref{sec:experiments} presents three sets of experiments: 1) a general comparison with other state-of-the-art network architectures for time-series forecasting, 2) an ablation study that highlights the effects of each proposed feature, and 3) an evaluation of online performance. Finally, Section \ref{sec:conclusions} concludes the limitations and possible future work.

\section{RELATED WORK}
\label{sec:background}
\subsection{Inertial Human Motion Capture}
\label{sec:human_motion_capture}
Inertial motion capture has been widely studied due to its advantages over optical-based systems, such as robustness to occlusions and unrestricted movement spaces. Commercial solutions \cite{xsens, noitom} employ dense IMU placement to achieve precise pose estimation, but this setup is intrusive and less practical for real-world applications. Prior research has attempted to reduce the number of required IMUs while maintaining accuracy. For instance, Marcard et al. used 6 IMUs for SMPL-based pose estimation \cite{von2017sparse}, while Huang et al. applied a learning-based method to infer poses from sparse IMU data \cite{Huang2018}. Yi et al. further introduced physics-based optimizations to improve motion plausibility \cite{Yi2021, Yi2022}, and Jiang et al. utilized a transformer-based model to enhance motion tracking \cite{Jiang2022}. However, these methods primarily focus on pose reconstruction rather than predicting future movements. Furthermore, they define motions in task space rather than configuration space, overlooking joint kinematics constraints that are important for physically feasible predictions. 

\subsection{Human Motion Prediction}
\label{sec:human_motion_prediction}
Human motion prediction is commonly formulated as a sequence-to-sequence learning problem, where future poses are inferred from historical poses. 
Martinez et al. proposed a simplified yet effective recurrent neural network (RNN) model for motion prediction, demonstrating promising performance \cite{martinez2017human}. Darvish et al. introduced a guided-mixture-of-experts approach to simultaneously recognize action and predict whole-body movements \cite{darvish2022simultaneous}. 
Gui et al. explored how robots can learn to anticipate human motion by analyzing past movement patterns \cite{gui2018teaching}. Meanwhile, Zhang et al. incorporated physics-based priors, leveraging estimated contact forces and joint torques to enhance motion prediction accuracy \cite{zhang2022pimnet}.
Beyond motion sequence models, some works integrate external sensor data to improve human motion forecasting. 
Yang et al. designed a recurrent model that estimates lower-body pose and foot contact states based on past upper-body signals \cite{Yang2021}. Zhang et al. introduced a two-stage motion forecasting network that predicts future human poses from historical IMU readings \cite{Zhang2023}.
However, these approaches often require extensive past pose data or complex multi-sensor fusion, which can limit their scalability to systems with a reduced set of sensors.

\begin{figure*}[th]
    \centering
    \includegraphics[width=0.9\textwidth]{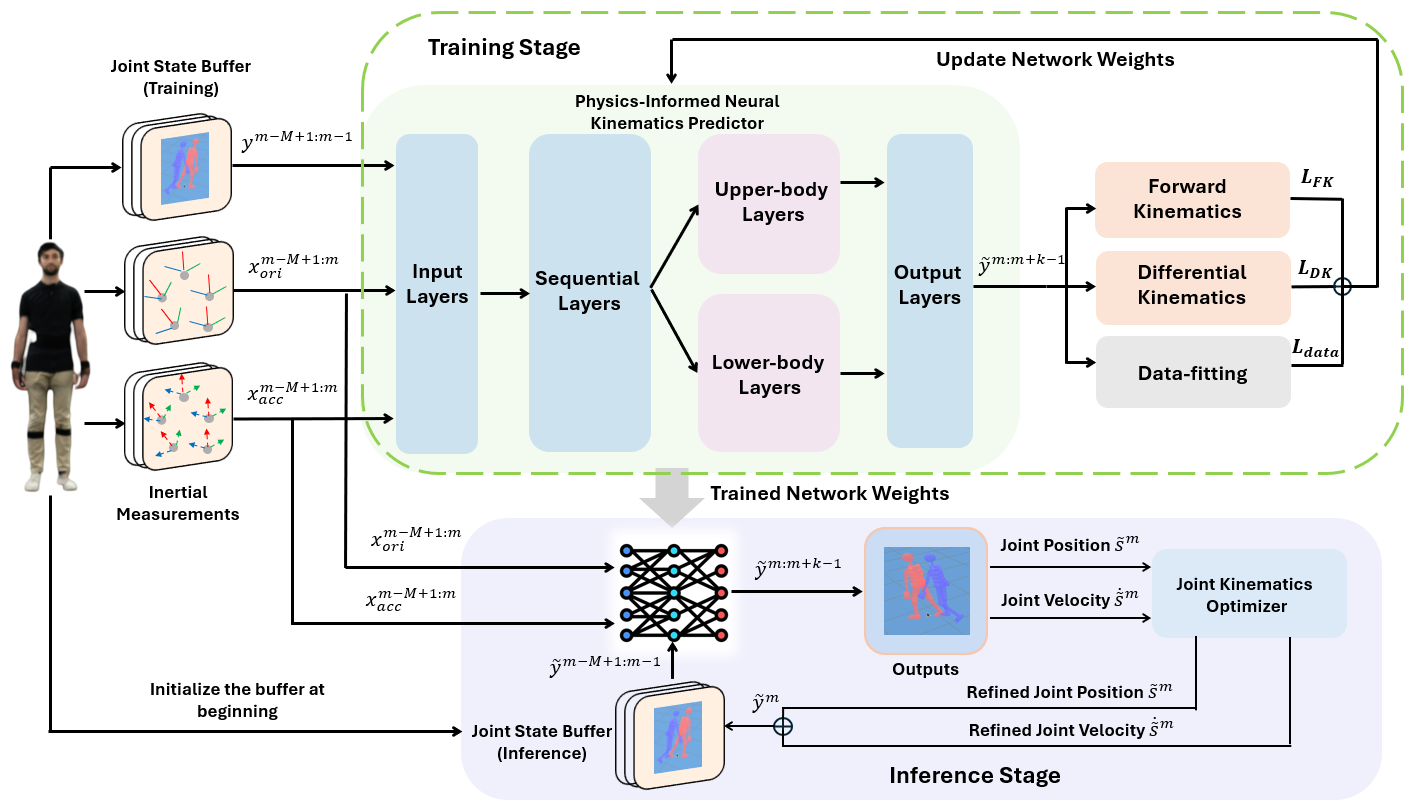}
    \caption{Overview of the proposed Physics-Informed Neural Kinematics Predictor. The green dashed box describes the training process. The network architecture is demonstrated within the light green area, while the \textit{physics-informed} and \textit{data-fit} losses are presented by side. The purple area at bottom displays the inference process, where the trained network is deployed. The joint state buffer is initialized with ground truth.}
    \label{fig:general_framework}
\end{figure*}
\subsection{PINN for Human Dynamics Modeling}
\label{sec:pinn}
Physics-Informed Neural Network was first proposed in \cite{Raissi2019} for solving supervised learning tasks that involve nonlinear partial differential equations. More recently, Zhang et al. expanded on this approach, presenting a physics-informed learning framework based on a convolutional neural network to predict synchronous muscle forces and joint kinematics from surface electromyogram (sEMG) signals \cite{Zhang2022}. Similarly, Shi et al. presented a physics-informed low-shot learning method for sEMG-based estimation of muscle forces and joint kinematics \cite{Shi2023}. They incorporated Lagrange's equations of motion and an inverse dynamics model into a generative adversarial network to enable structured feature decoding from limited sample data. In another work, Ma et al. embedded Hill muscle forward dynamics into the deep neural network to predict muscle forces without any label information during training \cite{Ma2024}. Nevertheless, current research is largely focused on sEMG-based methods, which are less practical for application in dynamic human-robot collaborative tasks.

\section{PROPOSED METHOD}
\label{sec:methods}
Figure \ref{fig:general_framework} exhibits an overview of the proposed method. Our goal is to predict future human motion in terms of whole-body joint kinematics, using a sequence of inertial readings from 5 IMUs mounted on the pelvis, both forearms, and both lower legs. 
Our novel approach, \emph{Physics-Informed Neural Kinematics Predictor} (PINKP), is trained in a supervised fashion by incorporating relevant physics knowledge alongside the data-fitting process. 
To further address the ambiguity arising from sparse IMU data, a joint state buffer is utilized as an additional input. 
The joint kinematics optimizer refines the previous first-step prediction to update the buffer, enabling a closed-loop autoregressive process. 

\subsection{Human Kinematics Modeling}
\label{sec:human_modeling}
The human is modeled as a floating-base multi-rigid-body dynamic system composed of \(n+1\) links and connected by \(n\) joints \cite{traversaro2017modelling}, 
whose configuration space lies on a Lie group composition \(\mathbb{Q}=\mathbb{R}^3 \times \text{SO}(3) \times \mathbb{R}^n\). 
An element from this space is represented as \(\textbf{q}=(\prescript{\mathcal{I}}{}{\mathbf{p}_{\mathcal{B}}}, \prescript{\mathcal{I}}{}{\mathbf{R}_{\mathcal{B}}},   \textbf{s})\), 
where \(\prescript{\mathcal{I}}{}{\mathbf{H}_{\mathcal{B}}}\) = (\(\prescript{\mathcal{I}}{}{\mathbf{p}_{\mathcal{B}}}\), \(\prescript{\mathcal{I}}{}{\mathbf{R}_{\mathcal{B}}}\)) indicates the pose of the base frame \(\mathcal{B}\) w.r.t. the Inertial frame \(\mathcal{I}\) and \(\textbf{s} \in \mathbb{R}^n\) being the joint positions that represent the system topology.
The system velocity belongs to the set \(\mathbb{V}=\mathbb{R}^3 \times \mathbb{R}^3 \times \mathbb{R}^n\),
an element of \(\mathbb{V}\) is given by 
\(\bm{\nu}\) = (\(\prescript{\mathcal{I}}{}{\mathbf{\dot{p}}_{\mathcal{B}}}\), \(\prescript{\mathcal{I}}{}{\bm{\omega}_{\mathcal{B}}}\), \(\dot{\textbf{s}}\)),
where 
\(\mathbf{v}_{\mathcal{B}}\) = \((\prescript{\mathcal{I}}{}{\mathbf{\dot{p}}_{\mathcal{B}}} , \prescript{\mathcal{I}}{}{\bm{\omega}_{\mathcal{B}}})\) 
being the twist of base frame w.r.t. \(\mathcal{I}\) and \(\dot{\textbf{s}}\) \(\in\) \(\mathbb{R}^n\) being the joint velocities. 
The \textit{forward kinematics} (FK) function \(\bm{\mathcal F}_i\) for a frame 
attached to link \textit{i} describes the mapping from system configuration \(\textbf{q}\) to link pose \(\prescript{\mathcal{I}}{}{\mathbf{H}_{i}}\):
\begin{equation} \label{forward_kinematics}
\bm{\mathcal F}_{i}: \mathbb{Q} \to \text{SE}(3), \mathbf{q} \mapsto \prescript{\mathcal{I}}{}{\mathbf{H}_{i}} = \bm{\mathcal F}_{i}(\mathbf{q}),
\end{equation}
we denote the operators for retrieving the link position \(\prescript{\mathcal{I}}{}{\mathbf{p}_{i}}\) and orientation \(\prescript{\mathcal{I}}{}{\mathbf{R}_{i}}\) as \(\bm{\mathcal F}_{i}^p\) and \(\bm{\mathcal F}_{i}^o\), respectively. 

The velocity of a frame 
attached to link \textit{i} is denoted as \begin{math}{{}\mathbf{v}_{i} = ({}^\mathcal{I}\mathbf{\dot{p}}_{i}, {}^\mathcal{I}\bm{\omega}_{i})}\end{math}.
The relationship between the \textit{i}-th link frame velocity \(\mathbf{v}_{i}\) and system configuration \((\mathbf{q} , \bm{\nu})\) is described by the \textit{differential kinematics} (DK) function \(\bm{\mathcal G}_{i}\):
\begin{equation} \label{jacobian}
\bm{\mathcal G}_{i}: \mathbb{Q} \times \mathbb{V} \to \mathbb{R}^{3} \times \mathbb{R}^{3}, \mathbf{q}, \bm{\nu} \mapsto \mathbf{v}_{i} = J_{i}(\mathbf{q}) \bm{\nu},
\end{equation}
where \(J_{i}(\mathbf{q})\) is the \textit{i}-th link \textit{Jacobian}.

\subsection{Problem Formulation}
\label{sec:problem_formulation}
Given a sequence of past inertial measurements 
\(X^{m-M+1:m} = [X^{m-M+1} , \dots , X^{m-1}, X^m] \in \mathbb{R}^{M \times N \times F}\)
(where \textit{m} is the current timestamp, \textit{M} is the input sequence length, \textit{N} is the number of IMUs, \textit{F} is the dimension of input features of each IMU),
we want to find a 
mapping function that can continuously predict the future joint configurations, namely 
\(\tilde{y}^{m:m+K-1} = f(\theta ; X^{m-M+1:m}) \in \mathbb{R}^{K \times 2 \times n}\) (where \textit{K} is the prediction horizon, \textit{n} denotes the dimension of joint degree of freedoms, $\theta$ indicates the hyperparameters of the mapping function).
More specifically, a single step inertial measurements \(X^i = [x_{acc}^i , x_{ori}^i] \in \mathbb{R}^{5 \times (3+9)}\) contains the acceleration and flattened rotation matrix
readings from all 5 IMUs,
while a single step prediction \(\tilde{y}^i = [\tilde{\mathbf{s}}^i, \dot{\tilde{\mathbf{s}}}^i] \in \mathbb{R}^{2 \times n}\) includes both joint position and velocity.

\subsection{Physics-Informed Neural Kinematics Predictor}
\label{sec:jk_predictor}
As illustrated in Figure \ref{fig:general_framework}, we use a time window of length \textit{M} to continuously collect inertial measurements of accelerations and orientations from all 5 selected links as inputs. 
However, a key challenge of using sparse IMUs is distinguishing between ambiguous poses when IMU readings are very similar. In this study, we assume that human movements are consistent and steady. Consequently, the prediction of future kinematics state \(\tilde{y}^{m:m+K-1}\) is conditioned on both sensor measurements \(X^{m-M+1:m}\) and previously estimated internal states \(\tilde{y}^{m-M+1:m-1}\), following the formulation \(\max P(\tilde{y}^{m:m+K-1} | \tilde{y}^{m-M+1:m-1}, X^{m-M+1:m})\). 
Therefore, we introduce a \textit{joint state buffer} of length (\(M-1\)) to retain historical joint kinematics, which serve as additional input to the network. During training, the joint history kinematics are obtained from the 
ground truth data. During inference, the initial joint configurations \(y^{0:M-2}\) are provided as known values, after which the buffer is updated recursively using the previous prediction, following a \textit{first-in-first-out} scheme.

\subsubsection{Network Architecture}
\label{sec:network}
The sequential inputs of inertial measurements \(X^{m-M+1:m}\) and joint history states \(\tilde{y}^{m-M+1:m-1}\) are initially processed through their respective fully connected layers utilizing the exponential linear unit (\textit{elu}) activation function to extract hidden temporal features. The resulting outputs are then concatenated and passed through two additional fully connected layers, also employing the \textit{elu} activation function. Recognizing that the dynamics of upper body and lower body motions can largely be considered independent, two separate blocks with identical structures are parallelly inserted to forecast the motion of the upper and lower bodies, respectively. Finally, the outputs from these two blocks are concatenated to facilitate the simulation of whole-body motions. 

\subsubsection{Loss Function Design}
\label{sec:loss}
To train the network, we proposed a loss function comprising two main components, namely, \textit{data-fit} loss and \textit{physics-informed} loss. The \textit{data-fit} loss 
\(L_\text{data}\)
is formulated as the Mean Squared Error (MSE) between the predicted and the target values:
\begin{equation} \label{L_data}
\begin{split}
L_\text{data} & = L_\text{pos} + L_\text{vel} \\
         & = \frac{1}{2K}\sum_{t=m}^{m+K-1}(\|\bm{\tilde{s}}^t-\bm{s}^t\|_2^2 + \|\dot{\tilde{\bm{s}}}^t - \dot{\bm{s}}^t\|_2^2),
\end{split}
\end{equation}
where $m$ is the current timestamp, \(\bm{s}\) and \(\dot{\bm{s}}\) are the targets.

Moreover, the predicted joint configuration should also 
comply with the physical constraints imposed by
Equations (\ref{forward_kinematics}) and (\ref{jacobian}), which compose the \textit{physics-informed} loss \(L_\text{phy}\):
\begin{subequations} \label{L_phy}
    \begin{align}
        L_\text{phy} &= L_\text{FK} + L_\text{DK} \label{eq:first},\\
        L_\text{FK} &= \frac{1}{DK}\sum_{i=1}^{D}\sum_{t=m}^{m+K-1} \bigg(
        \|\bm{\mathcal F}_{i}^{p}(\mathbf{\tilde{q}}^t) - \mathbf{p}_{i}^t\|_2^2  +\notag \\
        &\quad \|\bm{\mathcal F}_{i}^{o}(\mathbf{\tilde{q}}^t) \ominus \mathbf{R}_{i}^t\|_2^2 \bigg) \label{eq:second},\\
        L_\text{DK} &= \frac{1}{DK}\sum_{i=1}^{D}\sum_{t=m}^{m+K-1}\|\bm{\mathcal G}_{i}(\mathbf{\tilde{q}}^t, \bm{\tilde{\nu}}^t)-\mathbf{v}_{i}^t\|_2^2\label{eq:third}
    \end{align}
\end{subequations}
where \textit{D} is the number of selected links, 
\textit{K} is the prediction steps, \textit{m} is the current timestamp.
\(\mathbf{\tilde{q}}^t\) consists of predicted joint position \(\bm{\tilde{s}}^t\), and ground truth of base pose (\(\mathbf{p}_{\mathcal{B}}^t, \mathbf{R}_{\mathcal{B}}^t\)), while
\(\bm{\tilde{\nu}}^t\) includes predicted joint velocity \(\dot{\tilde{\bm{s}}}^t\), and ground truth of base twist \((\mathbf{\dot{p}}_{\mathcal{B}}^t, \bm{\omega}_{\mathcal{B}}^t)\). The values of \(\mathbf{\tilde{q}}^t\) and \(\bm{\tilde{\nu}}^t\) are utilized to obtain the estimated link pose and velocity at timestamp \textit{t}.
In the meanwhile, \(\mathbf{p}_{i}^{t}\), \(\mathbf{R}_{i}^{t}\) and \(\mathbf{v}_{i}^{t}\) denote the reference position, orientation and velocity of \textit{i}-th link frame at the same timestamp. 
The symbol $\ominus$ is for calculating the distances between the orientations, which are represented as flattened rotation matrix.  

The total loss involved to update the network is a weighted sum of each \textit{data-fit} and \textit{physics-informed} individual:
\begin{equation}\label{L_total}
L_\text{total} = \lambda_1 L_\text{pos} + \lambda_2 L_\text{vel} + \lambda_3 L_\text{FK} + \lambda_4 L_\text{DK},  
\end{equation}
where \(\lambda_i\), $i \in [1, 4]$ denote hyperparameters that are manually tuned to scale the significance of each loss component at the training stage.

\subsection{Joint Kinematics Optimizer}
\label{sec:buffer_update}
At the inference stage, the prediction errors persist and can accumulate due to the long-term drift and sensor noise. To address this, we introduce a joint kinematics optimizer that refines the network's prediction at timestamp \textit{m}, which is then used to update the \textit{joint state buffer}.
The refinement process is carried out by solving the following optimization problem:
\begin{align}
    \min_{\mathbf{s}^*, \mathbf{\dot{s}}^*} \quad & \|\bm{s}^*-\bm{\tilde{s}}^t\|^2_2+\|\dot{\bm{s}}^*-\dot{\tilde{\bm{s}}}^t\|_2^2 \label{eq:objective},\\
    \text{s.t.} \quad & \|\bm{\mathcal G}_{i}(\bm{s}^*, \mathbf{H}_{\mathcal{B}}^t, \dot{\bm{s}}^*, \mathbf{v}_{\mathcal{B}}^t)-\bm{v}^{t}_{i}\|^2_2 \leq \varepsilon, \quad i = 1, \dots, D \label{eq:constraint1}
\end{align}
where \(t=m\) indicates the current timestamp. Note that we decompose the original arguments \((\mathbf{q}^t, \bm{\nu}^t)\) for function \(\bm{\mathcal G}_{i}\) as \((\bm{s}^*, \mathbf{H}_{\mathcal{B}}^t, \dot{\bm{s}}^*, \mathbf{v}_{\mathcal{B}}^t)\) in the constraint. Our goal is to find a pair of \(\mathbf{s}^*, \mathbf{\dot{s}}^*\) that corrects potential prediction drift by enforcing consistency with the \textit{differential kinematics} equation for each selected link \textit{i}, while staying close to the initial guess provided by the network at timestamp \textit{m}. Importantly, only this single-step prediction is refined and stored in the \textit{joint state buffer}, ensuring that future predictions from step \(m+1\) to \(m+K-1\) are implicitly adjusted over time. 
We opted not to include \textit{forward kinematics} as a constraint because it serves a similar role to \textit{differential kinematics} but only constrains joint position predictions, potentially introducing unnecessary computational overhead.

\section{EXPERIMENTS AND RESULTS}
\label{sec:experiments}
\subsection{Experimental Setup}
\label{sec:setup}
\textbf{Dataset.} To the best of our knowledge, 
there is no publicly available dataset that includes both inertial measurements and whole-body human joint kinematics.
To address this, we leverage our CLIK-based framework \footnote{\href{https://github.com/ami-iit/bipedal-locomotion-framework}{$https://github.com/ami \_ iit/bipedal \_ locomotion \_ framework$}}, which processes data from an Xsens \cite{xsens} wearable sensing system equipped with 17 wireless IMUs distributed across the entire body. Our self-collected dataset captures a diverse range of whole-body motions performed at varying speeds on flat terrain, with continuously changing facing direction. Each recorded sequence includes stops and restarts. As summarized in Table \ref{tab:mocap_data}, each motion lasts approximately 8 min, resulting in a final dataset of about 1 hour, sampled at 60 Hz.

\newcommand{\ra}[1]{\renewcommand{\arraystretch}{#1}}
\begin{table}[]
\centering
\caption{Description of human motion dataset.}
\label{tab:mocap_data}
\resizebox{\columnwidth}{!}{%
\begin{tabular}{@{}llllllll@{}}
\toprule
\textbf{Motion Type} &
  \begin{tabular}[c]{@{}l@{}}Forward \\ Walking\end{tabular} &
  \begin{tabular}[c]{@{}l@{}}Forward \\ Walking\end{tabular} &
  \begin{tabular}[c]{@{}l@{}}Backward\\ Walking\end{tabular} &
  \begin{tabular}[c]{@{}l@{}}Side \\ Stepping\end{tabular} &
  \begin{tabular}[c]{@{}l@{}}Forward \\ Walking\\ Lifting Arms\end{tabular} &
  \begin{tabular}[c]{@{}l@{}}Forward\\ Walking\\ Waving Arms\end{tabular} &
  \begin{tabular}[c]{@{}l@{}}Forward\\ Walking \\ Clapping Hands\end{tabular} \\ \midrule
\textbf{Speed}              & Normal & Fast  & Normal & Normal & Normal & Normal & Normal \\
\textbf{Duration {[}min{]}} & 8.0    & 7.9   & 8.1    & 7.9    & 7.9    & 7.9    & 7.9    \\
\textbf{Frames}             & 28798  & 28332 & 29117  & 28440  & 29401  & 28508  & 28505  \\ \bottomrule
\end{tabular}%
}
\end{table}

\textbf{Metrics.} 
We define two metrics for the general evaluation of models' predictions, namely, \textit{Mean Absolute Joint Position Error} (pMAE) and \textit{Mean Absolute Joint Velocity Error} (vMAE) which measure the average absolute error in the predicted joint positions/velocities for selected Degrees of Freedom (DoFs). To further distinguish the effects of the physics-informed component, we utilize three more metrics: \textit{Root Mean Squared Link Position Error} (pRMSE), \textit{Root Mean Squared Link Orientation Error} (oRMSE) and \textit{Root Mean Squared Link Velocity Error} (vRMSE) that measure the root mean squared error of retrieved link position/orientation/velocity using the joint position predictions.

\textbf{Implementation Details.} The training and testing are conducted on a laptop with an Intel(R) Core(TM) i7-10750H CPU and an NVIDIA GeForce GTX 1650 Ti GPU. We use PyTorch with CUDA 12.2 to implement the kinematics prediction network and leverage the Automatic Differentiation Rigid-Body-Dynamics Algorithms Library \footnote{\href{https://github.com/ami-iit/adam}{$https://github.com/ami \_ iit/adam$}} to 
enable the differentiation of inserted physics-informed quantities during backpropagation. 
The network is trained for 30 epochs with a batch size of 256 using the Adam optimizer. 
The learning rate is initialized as 1e-3 and is updated by the StepLR scheduler every 5 epochs. 
The inertial sequence length that is fed to the network is equal to 10 steps, while the prediction horizon is set as 60 steps.

\subsection{Comparison Studies}
\label{sec:comparison}
We conduct a comparison of our method with several leading sequence forecasting networks: LSTM \cite{hochreiter1997long} (a canonical recurrent architecture), TCN \cite{bai2018empirical} (a temporal convolutional network for sequence modeling), and TIP \cite{Jiang2022} (a Transformer-based approach for human motion reconstruction). This comparison is performed across all tasks outlined in Table \ref{tab:mocap_data}. For comparative analysis, the inputs and outputs of the baseline networks have been adjusted to match our model, as depicted in Figure \ref{fig:general_framework}. 

\textbf{Quantitative.} In Table \ref{tab:quantitative_comparison}, we present a quantitative comparison of our method against baselines across various activities, each evaluated at different prediction timestamps (i.e., t=1, 30, 60). For the \textit{locomotion tasks} (first four activities), metrics are computed as the average errors of lower-body joints, while for the \textit{whole-body tasks} (last three activities), the analysis considers full-body joints. Our method consistently achieves lower mean absolute errors in predicting both joint positions and velocities across most activities and prediction timestamps compared to LSTM, TCN, and TIP, indicating superior predictive accuracy on joint position and velocity. Additionally, as shown in Table \ref{tab:infer_time}, our method 
requires on average less than 1 ms for each iteration inference, 
making it well-suited for real-time applications. 

We attribute this superior performance to the integration of physical priors 
at multiple levels. Therefore the proposed approach enhance the accuracy of joint position and velocity prediction along with smoothness of transitions between motions.
The physical constraints applied during training guide the prediction of joint states, which results in a less noisy initial guess for the joint state buffer. 
Then, during inference, the joint kinematics optimizer leverages inertial measurements from the selected links to further reduce the potential accumulation of prediction errors.
The combination of these techniques yields more accurate predictions and smoother motion transitions.
\begin{table*}[t]
\centering
\caption{Quantitative comparison with human kinematics prediction baselines.}
\label{tab:quantitative_comparison}
\resizebox{\textwidth}{!}{%
\begin{tabular}{@{}llllllllllllllll@{}}
\toprule
\multirow{2}{*}{\textbf{\begin{tabular}[c]{@{}l@{}}Prediction\\ Timestamp\end{tabular}}} & \multirow{2}{*}{\textbf{Method}} & \multicolumn{2}{l}{\textbf{\begin{tabular}[c]{@{}l@{}}Forward \\ Walking Normal\end{tabular}}} & \multicolumn{2}{l}{\textbf{\begin{tabular}[c]{@{}l@{}}Forward\\ Walking Fast\end{tabular}}} & \multicolumn{2}{l}{\textbf{\begin{tabular}[c]{@{}l@{}}Backward\\ Walking Normal\end{tabular}}} & \multicolumn{2}{l}{\textbf{\begin{tabular}[c]{@{}l@{}}Side-stepping\\ Normal\end{tabular}}} & \multicolumn{2}{l}{\textbf{\begin{tabular}[c]{@{}l@{}}Forward Walking\\ Lifting Arms\end{tabular}}} & \multicolumn{2}{l}{\textbf{\begin{tabular}[c]{@{}l@{}}Forward Walking\\ Waving Arms\end{tabular}}} & \multicolumn{2}{l}{\textbf{\begin{tabular}[c]{@{}l@{}}Forward Walking\\ Clapping Hands\end{tabular}}} \\ \cmidrule(l){3-16} 
 &  & \begin{tabular}[c]{@{}l@{}}pMAE\\ (deg)\end{tabular} & \begin{tabular}[c]{@{}l@{}}vMAE\\ (deg/s)\end{tabular} & \begin{tabular}[c]{@{}l@{}}pMAE\\ (deg)\end{tabular} & \begin{tabular}[c]{@{}l@{}}vMAE\\ (deg/s)\end{tabular} & \begin{tabular}[c]{@{}l@{}}pMAE\\ (deg)\end{tabular} & \begin{tabular}[c]{@{}l@{}}vMAE\\ (deg/s)\end{tabular} & \begin{tabular}[c]{@{}l@{}}pMAE\\ (deg)\end{tabular} & \begin{tabular}[c]{@{}l@{}}vMAE\\ (deg/s)\end{tabular} & \begin{tabular}[c]{@{}l@{}}pMAE\\ (deg)\end{tabular} & \begin{tabular}[c]{@{}l@{}}vMAE\\ (deg/s)\end{tabular} & \begin{tabular}[c]{@{}l@{}}pMAE\\ (deg)\end{tabular} & \begin{tabular}[c]{@{}l@{}}vMAE\\ (deg/s)\end{tabular} & \begin{tabular}[c]{@{}l@{}}pMAE\\ (deg)\end{tabular} & \begin{tabular}[c]{@{}l@{}}vMAE\\ (deg/s)\end{tabular} \\ \midrule
\multirow{4}{*}{t=1} & LSTM & 6.25 & 45.28 & 5.14 & 41.40 & 3.49 & 23.49 & 3.99 & 25.47 & 6.01 & 38.25 & 6.49 & 37.26 & 4.20 & 29.41 \\
 & TIP & 5.38 & 41.85 & 6.47 & 54.90 & 3.99 & 30.91 & 3.24 & 28.25 & 7.26 & 53.04 & 7.22 & 54.39 & 9.07 & 73.87 \\
 & TCN & 3.76 & 27.19 & \textbf{2.72} & 18.77 & 3.02 & 22.38 & 3.46 & 25.82 & 6.34 & 34.34 & 6.48 & 29.02 & 4.51 & 26.74 \\ \cmidrule(l){2-16} 
 & Ours & \textbf{2.11} & \textbf{12.58} & 3.80 & \textbf{15.18} & \textbf{2.20} & \textbf{14.09} & \textbf{2.41} & \textbf{19.93} & \textbf{5.24} & \textbf{28.35} & \textbf{2.73} & \textbf{14.83} & \textbf{3.53} & \textbf{17.93} \\ \midrule
\multirow{4}{*}{t=30} & LSTM & 6.22 & 45.52 & 4.77 & 36.87 & 3.41 & 22.56 & 4.01 & 25.32 & 6.56 & 35.89 & 6.89 & 37.73 & 4.57 & 31.75 \\
 & TIP & 5.46 & 39.25 & 6.25 & 52.69 & 4.08 & 29.22 & 3.33 & 28.57 & 8.33 & 47.10 & 8.44 & 43.34 & 8.43 & 79.77 \\
 & TCN & 3.84 & 27.42 & 2.77 & 19.29 & 3.11 & 21.42 & 3.57 & 27.05 & 6.47 & 36.45 & 6.56 & 29.91 & 4.76 & 30.19 \\ \cmidrule(l){2-16} 
 & Ours & \textbf{2.27} & \textbf{14.89} & \textbf{2.33} & \textbf{15.49} & \textbf{2.34} & \textbf{15.21} & \textbf{2.89} & \textbf{22.18} & \textbf{6.30} & \textbf{29.48} & \textbf{2.96} & \textbf{16.93} & \textbf{3.77} & \textbf{22.25} \\ \midrule
\multirow{4}{*}{t=60} & LSTM & 6.05 & 45.60 & 4.98 & 35.68 & 3.40 & 22.74 & 4.03 & 25.13 & 7.48 & 28.38 & 7.88 & 31.14 & 5.04 & 33.95 \\
 & TIP & 5.26 & 38.95 & 6.12 & 46.38 & 4.07 & 28.43 & 3.52 & 28.80 & 9.16 & 38.70 & 9.35 & 44.43 & 9.47 & 61.57 \\
 & TCN & 3.91 & 28.27 & 3.07 & 21.79 & 3.16 & 20.74 & 3.59 & 26.77 & \textbf{7.01} & 37.31 & 6.59 & 26.22 & 5.45 & 34.17 \\ \cmidrule(l){2-16} 
 & Ours & \textbf{2.63} & \textbf{18.47} & \textbf{2.71} & \textbf{18.68} & \textbf{2.60} & \textbf{17.84} & \textbf{3.17} & \textbf{24.16} & 7.22 & \textbf{27.34} & \textbf{3.60} & \textbf{19.44} & \textbf{4.29} & \textbf{28.05} \\ \bottomrule
\end{tabular}%
}
\end{table*}

\begin{table}[t]
\centering
\caption{Inference time (ms) per iteration of different methods on various tasks.}
\label{tab:infer_time}
\resizebox{\columnwidth}{!}{%
\begin{tabular}{@{}llllllll@{}}
\toprule
\textbf{Method} & \textbf{\begin{tabular}[c]{@{}l@{}}Forward \\ Walking \\ Normal\end{tabular}} & \textbf{\begin{tabular}[c]{@{}l@{}}Forward\\ Walking \\ Fast\end{tabular}} & \textbf{\begin{tabular}[c]{@{}l@{}}Backward\\ Walking \\ Normal\end{tabular}} & \textbf{\begin{tabular}[c]{@{}l@{}}Side\\ stepping\\ Normal\end{tabular}} & \textbf{\begin{tabular}[c]{@{}l@{}}Forward \\ Walking\\ Lifting \\ Arms\end{tabular}} & \textbf{\begin{tabular}[c]{@{}l@{}}Forward \\ Walking\\ Waving \\ Arms\end{tabular}} & \textbf{\begin{tabular}[c]{@{}l@{}}Forward \\ Walking\\ Clapping \\ Hands\end{tabular}} \\ \midrule
LSTM & \textbf{0.84} & \textbf{0.89} & \textbf{0.84} & \textbf{0.88} & \textbf{0.86} & \textbf{0.83} & \textbf{0.92} \\
TIP & 7.24 & 7.34 & 7.23 & 7.48 & 7.50 & 7.44 & 7.69 \\
TCN & 2.16 & 2.28 & 2.16 & 2.23 & 2.18 & 2.11 & 2.35 \\ \midrule
Ours & \textbf{0.86} & \textbf{0.95} & \textbf{0.89} & \textbf{0.94} & \textbf{1.11} & \textbf{1.12} & \textbf{1.22} \\ \bottomrule
\end{tabular}%
}
\end{table}

\begin{figure}[t]
    \centering
    \includegraphics[width=\linewidth]{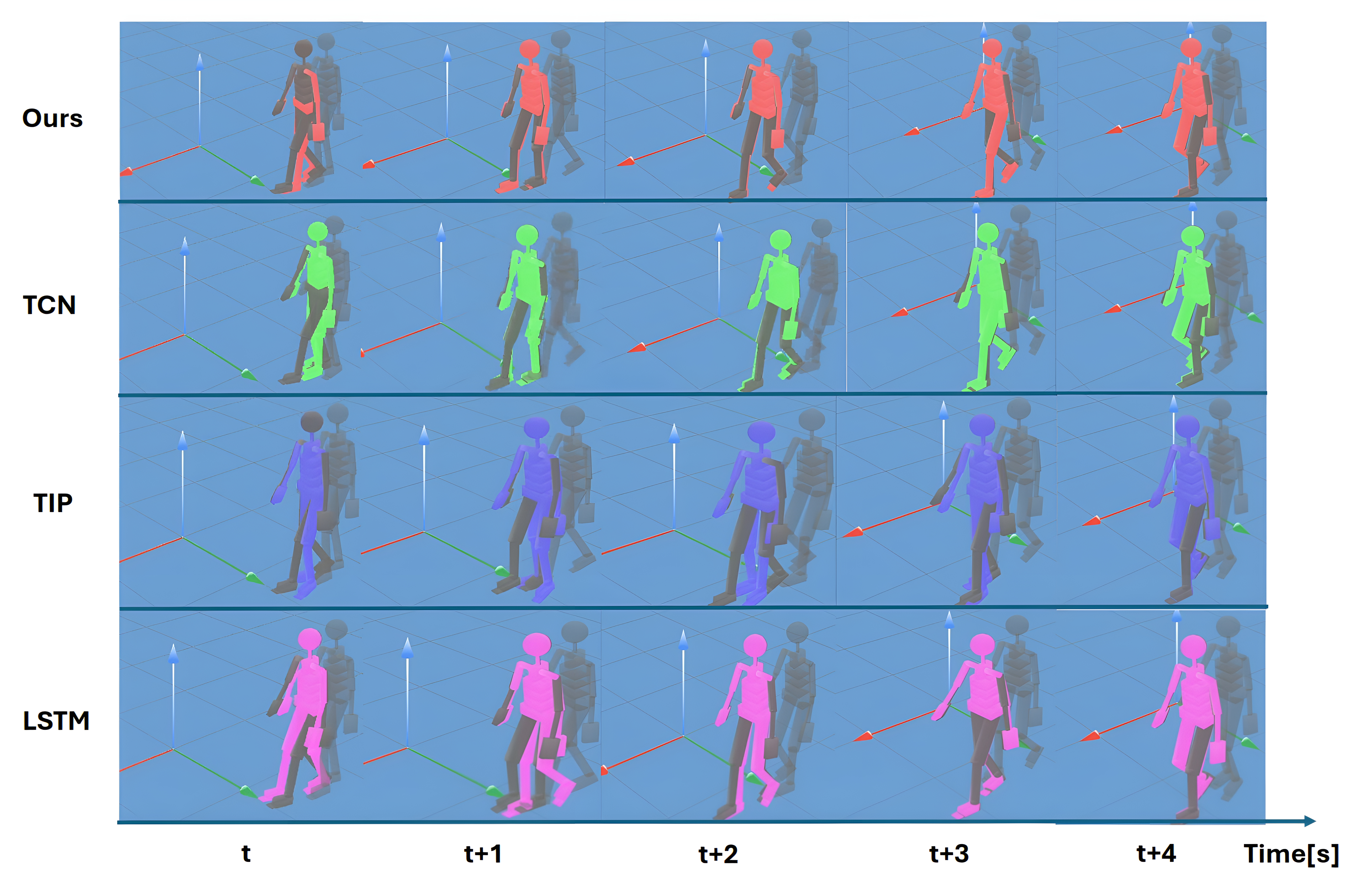}
    \caption{Qualitative results of different methods on the task \textit{forward walking with normal speed}. Screenshots are captured at 1 second intervals.}
    \label{fig:qualitative_plot}
\end{figure}

\textbf{Qualitative.} We further illustrate the qualitative comparison results in Figure \ref{fig:qualitative_plot}. The presented frames are captured from a sequence where the subject walks forward at an average pace. In each subplot, the semi-transparent gray avatar denotes the ground truth at the present timestamp. Meanwhile, the solid gray and colored avatars represent the ground truth and the prediction, projected 30 steps ahead, equivalent to 0.5 seconds into the future. As shown in the first column, our method 
can predict the subject's transition from a \textit{standing} pose to a \textit{walking} pose with mitigated motion delay thanks to the joint state buffer updated with refined prediction. In contrast, the baselines have difficulty precisely reconstructing the motion transition. Moreover, in the subsequent frames, our method demonstrates enhanced overlap with the reference, indicating higher predictive accuracy. 

\subsection{Ablation Studies}
\label{sec:ablation}
\textbf{Physics-Informed Component.}
In this section, we want to understand the impact of \emph{forward kinematics} (FK) and \emph{differential kinematics} (DK) on prediction performance. To isolate their effects, we disabled the joint state buffer, eliminating potential confounding influences from other modules. We trained four models, each with different physical loss components enabled: (1) no kinematics, (2) forward kinematics only, (3) differential kinematics only, and (4) both forward and differential kinematics. 

\begin{figure}[tb]
    \centering
    \begin{subfigure}[b]{0.5\textwidth}
        \centering
        \includegraphics[width=\linewidth]{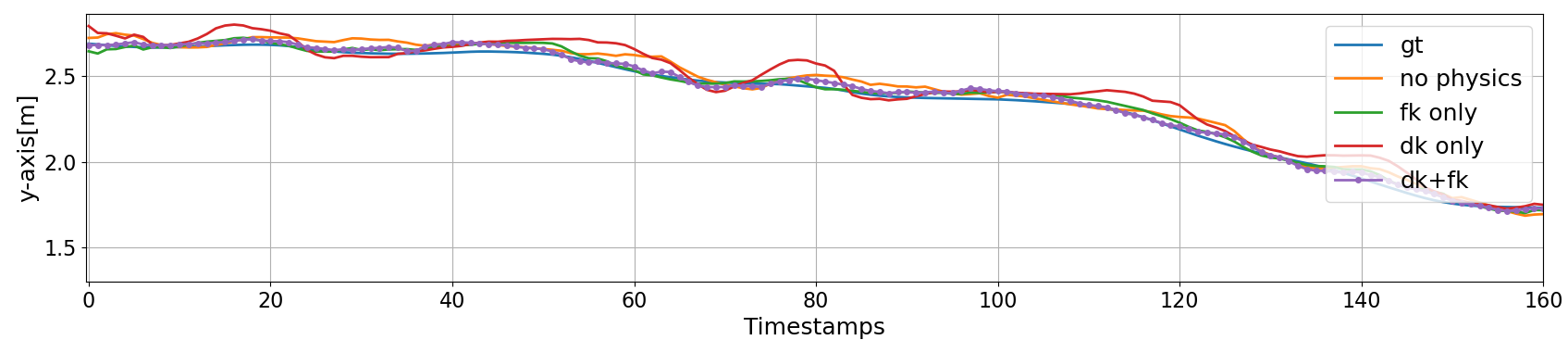}
        \caption{The retrieved link position along the y-axis.}
        \label{fig:subfig1}
    \end{subfigure}
    
    
    \begin{subfigure}[b]{0.5\textwidth}
        \centering
        \includegraphics[width=\linewidth]{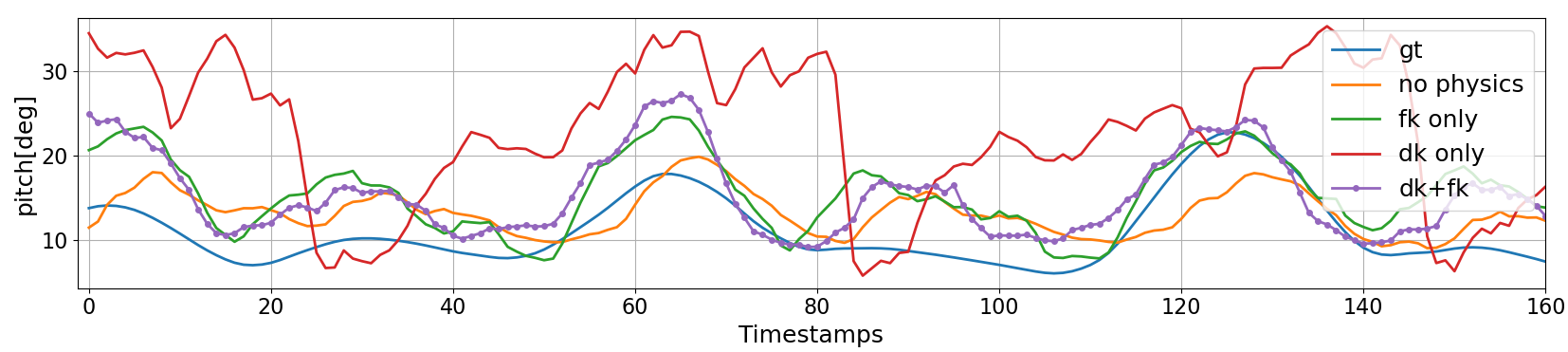}
        \caption{The retrieved pitch values (flexion/extension) of link orientation.}
        \label{fig:subfig2}
    \end{subfigure}

    \caption{A sequence of \textit{side-stepping} trajectory. The poses of the left lower leg are retrieved using first-step predictions from different models. The base is assumed to be given.}
    \label{fig:pi_fig}
\end{figure}

The four models were evaluated on a sequence of the \textit{Side Stepping} task. By leveraging the Equations (\ref{forward_kinematics}) and (\ref{jacobian}), we retrieved the poses and velocities of the desired link (e.g., left lower leg) using the 
first step of the sequences predicted by the ablation models.
The link position along the y-axis and the pitch values are illustrated in Figure \ref{fig:pi_fig}, while Table \ref{tab:pi_table} presents more detailed numerical results. 
We can see that different formulations of physical constraints influence the model's predictive performance in distinct ways. For instance, the model with only DK loss activated demonstrated strong accuracy in predicting link twist but performed worse in retrieving link poses compared to other models. In contrast, the model utilizing both FK and DK losses achieved the highest accuracy in terms of link positions and velocities but exhibited slightly reduced performance in link orientation compared to the baseline model that doesn't include the kinematics information.
We attribute this to the influence of the injected PI components, which incorporate physical principles that are not explicitly represented in the training data. By embedding the fundamental laws into the network, the PI components help the model generalize better, especially in scenarios with limited data.
\begin{table}[t]
\centering
\caption{Ablation study on the impact of different physics-informed formulations.}
\label{tab:pi_table}
\resizebox{\columnwidth}{!}{%
\begin{tabular}{@{}lllll@{}}
\toprule
\multirow{2}{*}{\textbf{Ablation Models}} & \multirow{2}{*}{\textbf{\begin{tabular}[c]{@{}l@{}}pRMSE\\ {[}m{]}\end{tabular}}} & \multirow{2}{*}{\textbf{\begin{tabular}[c]{@{}l@{}}oRMSE\\ {[}deg{]}\end{tabular}}} & \multicolumn{2}{l}{\textbf{vRMSE}} \\ \cmidrule(l){4-5} 
 &  &  & \textbf{\begin{tabular}[c]{@{}l@{}}Linear\\ {[}m/s{]}\end{tabular}} & \textbf{\begin{tabular}[c]{@{}l@{}}Angular\\ {[}deg/s{]}\end{tabular}} \\ \midrule
no kinematics & \textbf{0.0217} & \textbf{4.34} & 0.195 & 38.469 \\
forward kinematics only & 0.0252 & 4.80 & 0.168 & 36.936 \\
differential kinematics only & 0.0236 & 8.86 & \textbf{0.122} & \textbf{31.363} \\ \midrule
both forward and differential kinematics  & \textbf{0.0195} & \textbf{4.50} & \textbf{0.118} & \textbf{34.637} \\ \bottomrule
\end{tabular}%
}
\end{table}

\textbf{Joint State Buffer.}
\begin{figure*}[th]
    \centering
    \begin{subfigure}{0.45\textwidth}
        \centering
        \includegraphics[width=\linewidth]{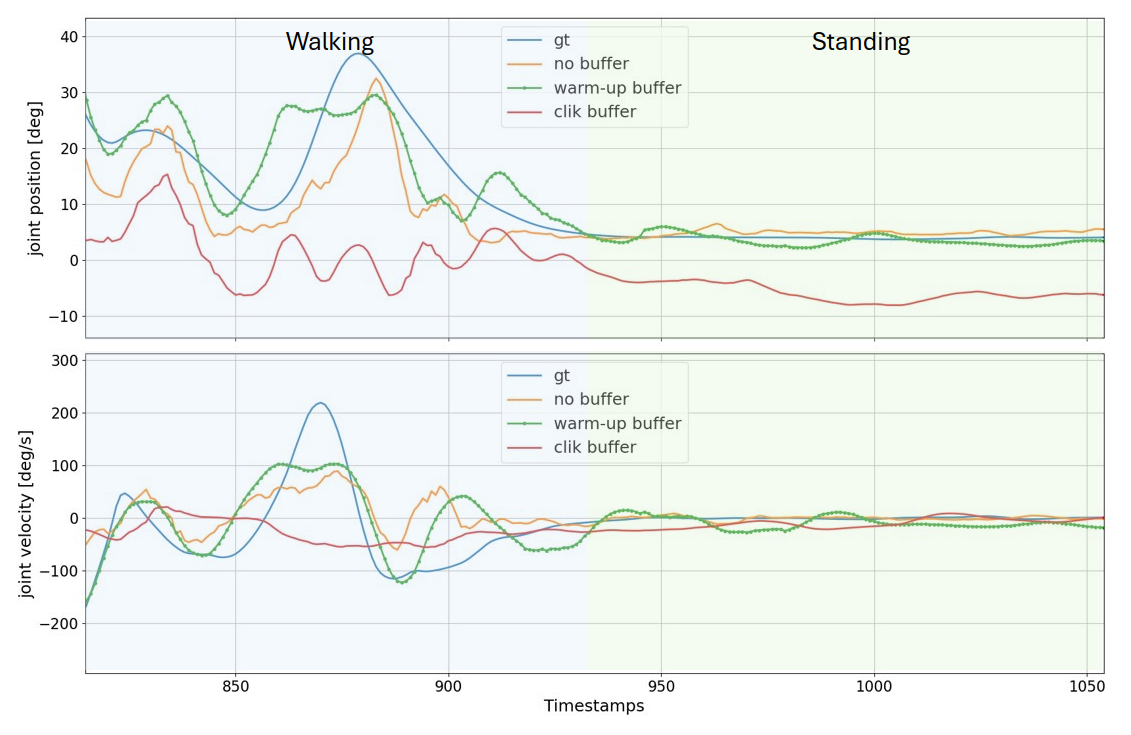}
        \caption{The transition from \textit{walking} to \textit{standing}.}
        \label{fig:fig1}
    \end{subfigure}
    \begin{subfigure}{0.45\textwidth}
        \centering
        \includegraphics[width=\linewidth]{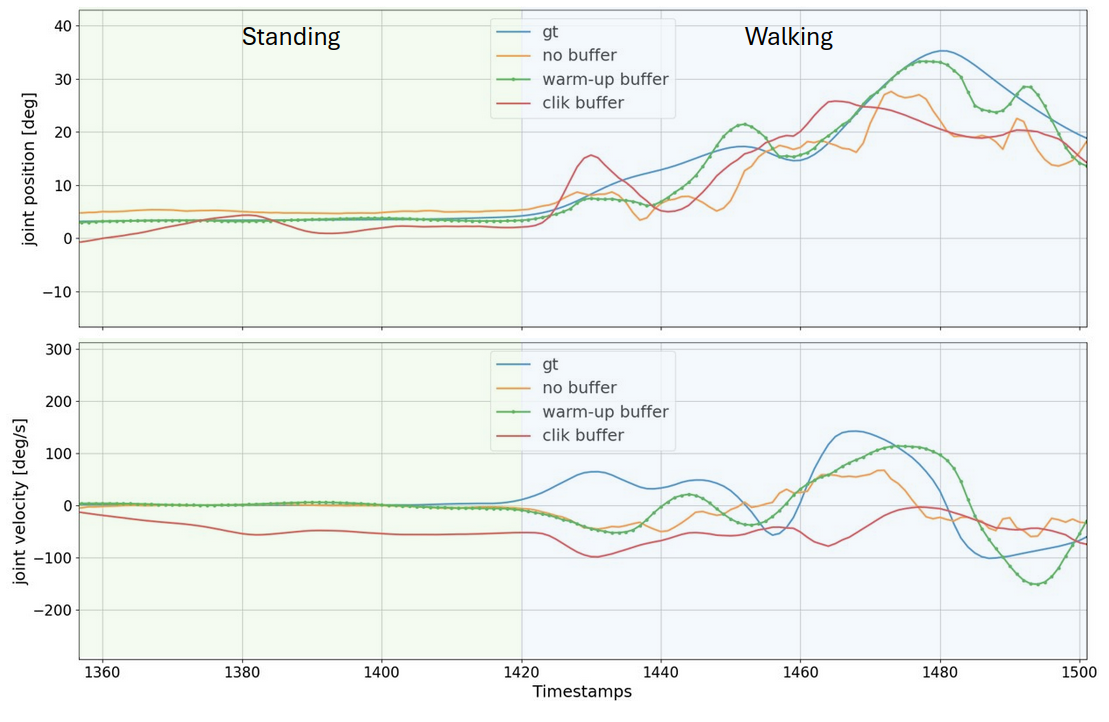}
        \caption{The transition from \textit{standing} to \textit{walking}.}
        \label{fig:fig2}
    \end{subfigure}
    \caption{A sequence of \textit{side-stepping} trajectory. The position and velocity of joint \textit{left knee} around the y-axis are compared since it represents the flexion and extension movement during walking. The blue line is the ground truth. The orange, red, and green lines indicate the first-step predictions of models 1 to 3, respectively.}
    \label{fig:buffer_fig}
\end{figure*}
\begin{figure*} 
\centering
\begin{subfigure}{0.85\textwidth}
    \includegraphics[width=\linewidth]{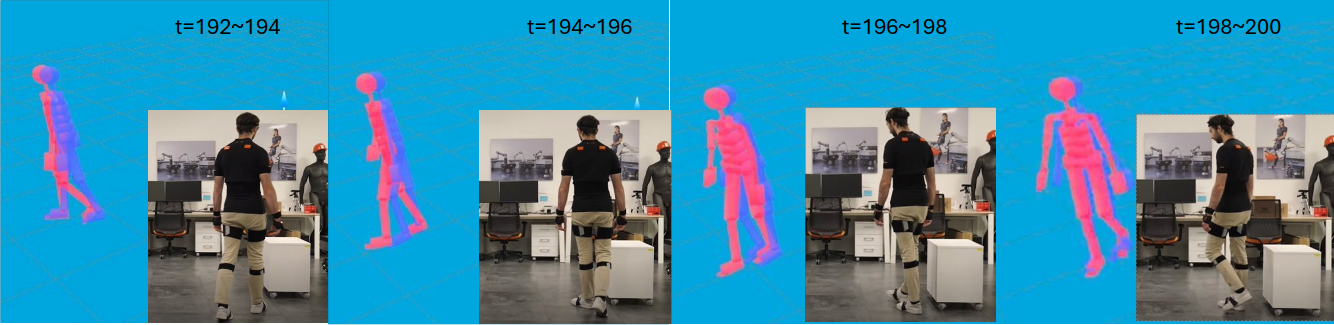}
    \caption{Recording of the volunteer walking around. The time displayer at the top (e.g., t=192 $\sim$ 194 sec) represents the interval during which the motion occurs. The analysis starts at 190 sec because the subject remains standing for a while at the beginning of each experiment. The blue avatar is the ground truth at the current timestamp, and the red avatar is the prediction at 10 timestamps ahead, sampled at 60 Hz.}
\end{subfigure}
\hfill
\begin{subfigure}{0.85\textwidth}
    \includegraphics[width=\linewidth]{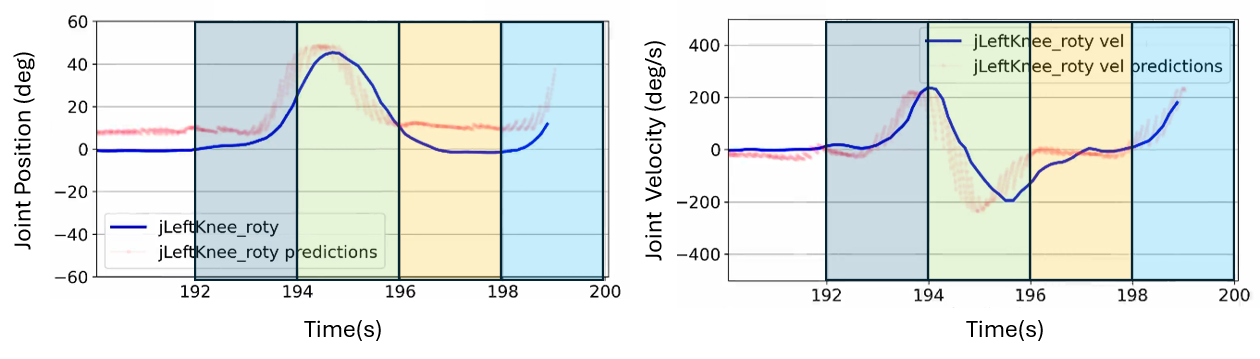}
    \caption{(Left) Joint position of left knee rotation around y-axis. (Right) Joint velocity of left knee rotation around y-axis. The blue line denotes the ground truth, while the red dotted line denotes the 10 timestamps prediction in the future.}
\end{subfigure}
\caption{Online experimental results of predicting human joint kinematics via the inertial readings of a sparse set of IMUs.}
\label{fig:online_test}
\end{figure*}
In this section, we examine the impact of the joint state buffer. We deployed three models, namely, (1) \textit{no buffer}: using only inertial inputs, (2) \textit{CLIK buffer}: incorporating both inertial inputs and historical states, with the state buffer updated via a CLIK-based IK module, that is, given inertial measurements of 5 IMUs, obtain the one-step joint configuration via a closed-loop iterative algorithm, and (3) \textit{warm-up buffer}: incorporating both inertial inputs and historical states, with the state buffer updated using the joint kinematics optimizer.
We tested the three ablation models on a sequence of \textit{Side Stepping} trajectory, which includes transitions between \textit{standing} and \textit{walking} modes. Figure \ref{fig:buffer_fig} illustrates the positions and velocities of the \textit{Left Knee} joint around the y-axis, capturing its flexion-extension movement. 
The third model, which utilizes the joint kinematics optimizer to update the buffer, showcased a much smoother transition between \textit{standing} and \textit{walking} poses compared to the other models. Additionally, it also provided more accurate predictions for both joint position and velocity when the subject remains stationary. The results of 
the second model, which leverages a CLIK-based module to update the buffer, suggests that updating the buffer solely through iterative IK solutions is suboptimal. Given that only five IMUs provide inertial measurements, the computed IK-based updates can be less reliable than directly refining the first-step prediction from the network. 

\subsection{Online Performance Evaluation}
\label{sec:online}
We evaluate our method using 17 Xsens IMUs for calibration and video recording, while processing only inertial data from five IMUs on the pelvis, forearms, and lower legs.
As depicted in Figure \ref{fig:online_test}, 
the subject begins with the right foot in contact, lifting the left foot at t = 193 s. The left knee’s velocity (y-axis) peaked first, followed by its maximum position. At t = 196.5 s, the right foot regained contact, and the left knee’s motion returned to zero. A new walking cycle started at t = 198 s. 
As illustrated in the third row, our method accurately captures the walking patterns, enabling precise predictions. Additionally, our method demonstrates its ability to generalize to the kinematics of unseen subjects.

\subsection{Limitation Discussion}
\label{sec:limitation}
Despite the promising performance in demonstrations, our work has several limitations. First, while our dataset includes diverse activities, subject diversity could be further improved. Second, our method assumes smooth and consistent motions, limiting its ability to handle discontinuous movements or rapid external perturbations. Lastly, relying on a predefined floating base (i.e., pose and velocity) constrains broader real-world applications. 
\section{CONCLUSIONS}
\label{sec:conclusions}
In this work, we present a physics-informed learning architecture to predict human kinematics using only 5 wearable IMUs. 
First, we represent the human motions in a constrained joint configuration space rather than the task space (i.e., limb poses). 
Then, we propose a network architecture that accounts for the temporal and spatial characteristics of human motion by separately modeling the upper and lower bodies within a unified framework.
By incorporating forward and differential kinematics as additional components alongside data-fit loss during training, the network 
is guided to learn the underlying relations between joints and limbs, which are not explicitly present in the training data.
At last, a joint kinematics optimizer is deployed at the inference stage to further enhance motion smoothness by updating a state buffer with a refined prediction.

In contrast to prior PINN-based approaches in biomechanics, our approach enables accurate, efficient full-body kinematics prediction with minimal sensors, eliminating the need for complex setups and fusion.
Concerning the current limitations, our future work will first focus on tracking the pose and velocity of the floating base. Expanding the dataset to include a more diverse range of subjects will also enhance our model generalization. 






\bibliographystyle{IEEEtran}
\bibliography{refs}

\end{document}